\documentclass[twoside,11pt]{article}

\usepackage[abbrvbib, preprint]{jmlr2e}

\usepackage{jmlr2e}

\usepackage{amsmath}
\usepackage{booktabs}
\usepackage{tablefootnote}
\usepackage{pifont}
\usepackage{threeparttable}
\usepackage{xcolor}
\usepackage{microtype}
\usepackage{enumitem}
\usepackage{listings}

\definecolor{codeblue}{RGB}{0,102,204}
\definecolor{codepurple}{RGB}{153,51,204}
\definecolor{codegreen}{RGB}{0,128,0}
\definecolor{codegray}{RGB}{128,128,128}
\definecolor{codebackground}{RGB}{248,248,248}
\definecolor{codeframe}{RGB}{220,220,220}

\renewcommand{\vec}[1]{\mbox{\boldmath${#1}$}}

\definecolor{xgreen}{RGB}{43,101,35}
\definecolor{xred}{RGB}{155,29,22}
\renewcommand{\checkmark}{\textcolor{xgreen}{ \ding{52}}}
\newcommand{\xmark}{\textcolor{xred}{\ding{56}}}

\jmlrheading{21}{2025}{1-8}{4/25}{5/25}{meila00a}{Selim F. Yilmaz and Suleyman S. Kozat}

\ShortHeadings{PySAD: A Streaming Anomaly Detection Framework in Python}{Yilmaz and Kozat}
\firstpageno{1}

\begin{document}

\title{PySAD: A Streaming Anomaly Detection Framework in Python}

\author{\name Selim F. Yilmaz \email s.yilmaz21@imperial.ac.uk \\
       \addr Department of Electrical and Electronic Engineering\\
       Imperial College London\\
       London, United Kingdom
       \AND
       \name Suleyman S. Kozat \email kozat@ee.bilkent.edu.tr \\
       \addr Department of Electrical and Electronic Engineering\\
       Bilkent University\\
       Ankara, Turkey}

\editor{}

\maketitle
\begin{abstract}
Streaming anomaly detection requires algorithms that operate under strict constraints: bounded memory, single-pass processing, and constant-time complexity. We present \textsf{PySAD}, a comprehensive Python framework addressing these challenges through a unified architecture. The framework implements 17+ streaming algorithms (LODA, Half-Space Trees, xStream) with specialized components including projectors, probability calibrators, and postprocessors. Unlike existing batch-focused frameworks, \textsf{PySAD} enables efficient real-time processing with bounded memory while maintaining compatibility with \textsf{PyOD} and \textsf{scikit-learn}. Supporting all learning paradigms for univariate and multivariate streams, \textsf{PySAD} provides the most comprehensive streaming anomaly detection toolkit in Python. The source code is publicly available at \href{https://github.com/selimfirat/pysad}{\texttt{github.com/selimfirat/pysad}}.
\end{abstract}
\section{Introduction}
Anomaly detection on streaming data has become critical in real-time analytics, driven by applications in cybersecurity~\citep{yuan2014online}, network intrusion~\citep{kloft2010online}, and face presentation attack detection~\citep{yilmaz2020face}. Modern data streams require algorithms that can process data points in real-time while adapting to evolving patterns and concept drift~\citep{gama2014survey}.

Streaming anomaly detection imposes stringent constraints: \textit{single-pass processing}, \textit{bounded memory usage}, \textit{constant-time processing}, and \textit{adaptive learning}. These constraints eliminate global optimization possibilities and require fundamentally different algorithmic approaches~\citep{henzinger1998computing}.

Existing frameworks reveal significant gaps: \textsf{scikit-learn}~\citep{sklearn} focuses on batch processing, \textsf{River}~\citep{montiel2021river} provides limited anomaly detection, and \textsf{PyOD}~\citep{pyod} lacks streaming optimizations. This fragmentation necessitates a dedicated streaming-focused framework.

We introduce \textsf{PySAD}, a comprehensive Python framework for streaming anomaly detection. The framework provides 17+ algorithms, from classical approaches (LODA~\citep{pevny2016loda}, Half-Space Trees~\citep{tan2011fast}) to modern ensemble methods (xStream~\citep{manzoor2018xstream}, sequential ensemble learning~\citep{yilmaz2020robust}), supporting univariate and multivariate streams across supervised, semi-supervised, and unsupervised paradigms~\citep{yilmaz2021unsupervised}.

Beyond core algorithms, \textsf{PySAD} provides a complete ecosystem: stream simulators, evaluation metrics, adaptive preprocessors, statistical trackers, probability calibrators, postprocessors, and batch-to-streaming integration utilities. The framework emphasizes production readiness through rigorous engineering practices and performance optimizations ensuring sub-millisecond processing.

The latest \textsf{PySAD} is available via \verb|pip install -U pysad| with comprehensive \href{https://pysad.readthedocs.io/}{\texttt{documentation}} and \href{https://github.com/selimfirat/pysad}{\texttt{examples}}.
\section{Streaming Anomaly Detection and \textsf{PySAD}}
\begin{figure}[t!]
    \centering
    \includegraphics[width=\textwidth]{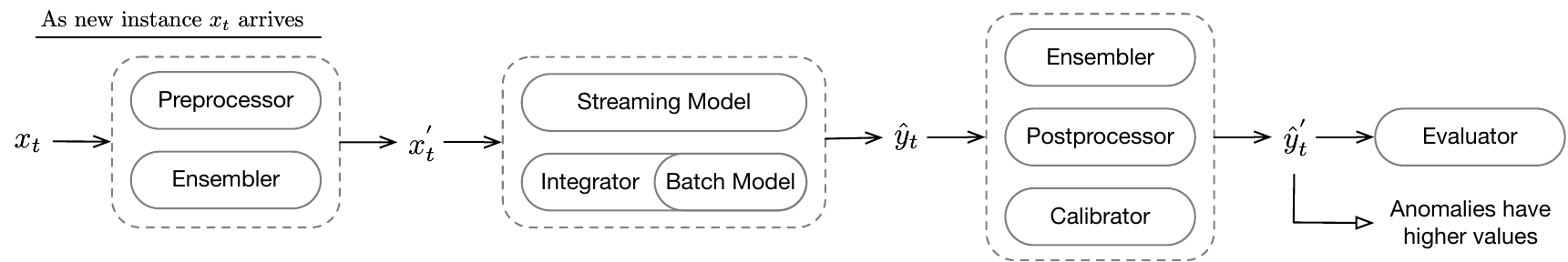}
    \caption{The usage of components in \textsf{PySAD} as a pipeline.}
    \label{fig:modules}
\end{figure}

A streaming anomaly detection model $\mathcal{M}$ receives a potentially infinite stream $\mathcal{D} = \{(\vec{x}_t, y_t) \mid t=1,2,...\}$, where $\vec{x}_t \in \mathbb{R}^m$ is a feature vector and $y_t$ is the binary anomaly label:
\begin{align*}
y_t =
\begin{cases}
1, & \text{ if } \vec{x}_t \text{ is anomalous,} \\
0, & \text{ otherwise.}
\end{cases}
\end{align*}

Streaming anomaly detection operates under three fundamental constraints: \textbf{single-pass processing} (each instance observed once), \textbf{bounded memory} (constant or sublinear growth), and \textbf{constant-time processing} (bounded per-instance complexity). These constraints eliminate traditional optimization approaches and necessitate online learning algorithms~\citep{gama2014survey}.

All models in \textsf{PySAD} extend the {\it BaseModel} class providing:
\begin{itemize}[nosep,leftmargin=10pt]
\item \texttt{fit\_partial($\vec{x}_t,\, y_t$)}: Incrementally trains using instance $(\vec{x}_t, y_t)$
\item \texttt{score\_partial($\vec{x}_t$)}: Returns anomaly score for $\vec{x}_t$
\item \texttt{fit\_score\_partial($\vec{x}_t, y_t$)}: Combines training and scoring
\end{itemize}

Figure~\ref{fig:modules} illustrates \textsf{PySAD}'s modular architecture:

{\bf Preprocessors} transform input data for normalization and scaling in streaming scenarios. {\bf Projectors} map data to lower-dimensional spaces for efficiency. {\bf Models} form the core detection component with classical (LODA, Half-Space Trees) and ensemble methods (xStream). {\bf Ensemblers} combine multiple model outputs and can be used to combine data points (early fusion) or decisions (late fusion)~\citep{mandira2019spatiotemporal,giritliouglu2021multimodal,yilmaz2020robust}. {\bf Postprocessors} refine scores through temporal smoothing and adaptive thresholding. {\bf Probability Calibrators} convert scores into interpretable probabilities using Gaussian tail fitting~\citep{ahmad2017unsupervised} or conformal prediction~\citep{ishimtsev2017conformal}. One can add models by extending {\it BaseModel} and implementing \texttt{fit\_partial} and \texttt{score\_partial}. Details are available at \href{https://pysad.readthedocs.io/}{\texttt{pysad.readthedocs.io}}.

\subsection{Usage Example}
The following example demonstrates typical \textsf{PySAD} usage for streaming anomaly detection:

\begin{lstlisting}
from pysad.evaluation.metrics import AUROCMetric
from pysad.models.loda import LODA
from pysad.utils.data import Data

model = LODA()
metric = AUROCMetric()
streaming_data = Data().get_iterator("arrhythmia.mat")

for x, y_true in streaming_data:
    anomaly_score = model.fit_score_partial(x)
    metric.update(y_true, anomaly_score)

print(f"Area under ROC metric is {metric.get()}.")
\end{lstlisting}

This example showcases the framework's simplicity: initialization requires minimal configuration, streaming data processing follows the standard \texttt{fit\_score\_partial} pattern, and evaluation metrics are updated incrementally.

\section{Comparison with Related Software}
Existing streaming anomaly detection frameworks can be categorized into (i) general streaming machine learning frameworks and (ii) batch-oriented anomaly detection libraries.

{\bf Streaming Frameworks:} \textsf{River}~\citep{montiel2021river} and \textsf{skmultiflow}~\citep{montiel2018scikit} implement only Half-Space Trees~\citep{tan2011fast} for streaming anomaly detection. \textsf{CapyMOA}~\citep{gomes2025capymoaefficientmachinelearning} provides 3 models through Python interfaces to MOA's Java algorithms~\citep{bifet2010moa}. \textsf{Jubat.us}~\citep{hido2013jubatus} implements only Local Outlier Factor~\citep{breunig2000lof} in C++. \textsf{Alibi-detect} offers limited streaming methods~\citep{ren2019time,le2005association}.

{\bf Batch Frameworks:} \textsf{PyOD}~\citep{pyod} and \textsf{ADTK}~\citep{adtk} excel in offline scenarios but lack streaming capabilities and do not address concept drift, memory constraints, or real-time processing requirements.

{\bf \textsf{PySAD}'s Position:} \textsf{PySAD} is purpose-built for streaming anomaly detection with 17+ specialized algorithms and uniquely provides unsupervised probability calibrators for converting raw scores into interpretable probabilities~\citep{safin2017conformal}.

Table~\ref{tab:comparisons} presents a comprehensive comparison highlighting \textsf{PySAD}'s distinctive focus on streaming anomaly detection and its comprehensive toolkit for building end-to-end streaming pipelines.

\begin{table}[t!]
\centering
\resizebox{\textwidth}{!}{
\begin{threeparttable}
\begin{tabular}{@{}lcccccccccc@{}}
\toprule
& \textsf{river} &  \textsf{jubat.us} &  \textsf{adtk} & \textsf{pyod} &  \textsf{skmultiflow}  & \textsf{alibi-detect} &  \textsf{moa} & \textsf{capymoa} & {\bf \textsf{pysad}}  \\ \midrule
Language & Python & C++  & Python & Python & Python &   Python & Java & Python &  Python  \\
\# of Models\footnotesize{*} & 5 & 1 & 0 & 0 & 1 & 2 & 6 & 3 & \textbf{17+} \\
Streaming & \checkmark & \checkmark & \xmark & \xmark & \checkmark & \checkmark & \checkmark & \checkmark & \checkmark  \\
Transformers & \checkmark & \checkmark  & \checkmark & \xmark & \checkmark &  \xmark  & \checkmark & \checkmark & \checkmark \\
Projectors & \checkmark & \xmark & \xmark & \xmark & \xmark & \xmark & \xmark & \xmark & \checkmark \\
Ensemblers & \checkmark & \xmark & \checkmark & \checkmark & \checkmark & \xmark & \checkmark & \xmark & \checkmark \\
Calibrators & \xmark & \xmark & \xmark & \xmark & \xmark & \xmark & \xmark & \xmark & \checkmark \\
\bottomrule
\end{tabular}
\begin{tablenotes}\footnotesize
\item[*] The number of specialized algorithms for streaming anomaly detection.
\item Current versions: \textsf{river} (0.22.0), \textsf{jubat.us} (1.1.1), \textsf{adtk} (0.6.2), \textsf{pyod} (2.0.5),  \textsf{skmultiflow} (0.5.3), \textsf{alibi-detect} (0.12.0), \textsf{moa} (24.07.0), \textsf{capymoa} (0.9.1), \textsf{pysad} (0.3.0).
\end{tablenotes}
\end{threeparttable}
}
\caption{Comparison with existing frameworks for streaming anomaly detection.}
\label{tab:comparisons}
\end{table}

As a specialized streaming anomaly detection framework, \textsf{PySAD} complements existing streaming frameworks and batch-oriented anomaly detection libraries while addressing the unique challenges of real-time anomaly detection.
\section{Development and Architecture}
\textsf{PySAD} is architected as a production-ready framework emphasizing scalability, maintainability, and performance. The framework is distributed under the BSD 3-Clause License for broad compatibility.

\subsection{Software Engineering Practices}
Our development methodology emphasizes quality assurance and collaborative development:

\begin{itemize}[nosep]
\item {\bf Collaborative Development:} Hosted on GitHub with issue tracking, pull request workflows, and active community contributions.

\item {\bf Quality Assurance:} 95\%+ code coverage, continuous integration across multiple platforms, PEP8 compliance, and comprehensive API documentation.

\item {\bf Performance Optimization:} Memory-efficient NumPy vectorization, constant-time algorithms, and sub-millisecond processing for high-throughput streams.

\item {\bf Minimal Dependencies:} Core dependencies include NumPy~\citep{walt2011numpy}, scikit-learn~\citep{sklearn}, SciPy~\citep{virtanen2020scipy}, and selective PyOD~\citep{pyod} integration.
\end{itemize}

\subsection{Architectural Design}
The framework implements modular architecture based on the Strategy pattern:

\begin{itemize}[nosep]
\item {\bf Interface Consistency:} Standardized interfaces (\texttt{BaseModel}, \texttt{BaseTransform}, \texttt{BaseMetric}) ensure seamless interoperability.

\item {\bf Memory Safety:} Automatic memory management with configurable bounds and leak prevention.

\item {\bf Extensibility:} Plugin architecture for easy algorithm contributions with minimal interface implementation.

\item {\bf Production Readiness:} Thread-safe implementations, comprehensive logging, and graceful error handling.
\end{itemize}

\textsf{PySAD} supports Python 3.10+ and installs via PyPI (\texttt{pip install pysad}) with automatic dependency resolution.


\acks{This work is supported by the Turkish Academy of Sciences Outstanding Researcher Programme and Tubitak Contract No: 117E153. We thank all contributors and the open-source community for their valuable feedback and contributions.}



\vskip 0.2in
\bibliography{ref}

\begin{thebibliography}{28}
\providecommand{\natexlab}[1]{#1}
\providecommand{\url}[1]{\texttt{#1}}
\expandafter\ifx\csname urlstyle\endcsname\relax
  \providecommand{\doi}[1]{doi: #1}\else
  \providecommand{\doi}{doi: \begingroup \urlstyle{rm}\Url}\fi

\bibitem[{A}hmad et~al.(2017){A}hmad, {L}avin, {P}urdy, and
  {A}gha]{ahmad2017unsupervised}
S.~{A}hmad, A.~{L}avin, S.~{P}urdy, and Z.~{A}gha.
\newblock {U}nsupervised real-time anomaly detection for streaming data.
\newblock \emph{{N}eurocomputing}, 262:\penalty0 134--147, 2017.

\bibitem[{B}ifet et~al.(2010){B}ifet, {H}olmes, {P}fahringer, {K}ranen,
  {K}remer, {J}ansen, and {S}eidl]{bifet2010moa}
A.~{B}ifet, G.~{H}olmes, B.~{P}fahringer, P.~{K}ranen, H.~{K}remer,
  T.~{J}ansen, and T.~{S}eidl.
\newblock {M}oa: {M}assive online analysis, a framework for stream
  classification and clustering.
\newblock In \emph{{P}roceedings of the {F}irst {W}orkshop on {A}pplications of
  {P}attern {A}nalysis}, pages 44--50, 2010.

\bibitem[{B}reunig et~al.(2000){B}reunig, {K}riegel, {N}g, and
  {S}ander]{breunig2000lof}
M.~M. {B}reunig, H.-P. {K}riegel, R.~T. {N}g, and J.~{S}ander.
\newblock {L}of: {I}dentifying density-based local outliers.
\newblock In \emph{{P}roceedings of the 2000 {A}{C}{M} {S}{I}{G}{M}{O}{D}
  {I}nternational {C}onference on {M}anagement of {D}ata}, pages 93--104, 2000.

\bibitem[Gama et~al.(2014)Gama, {\v{Z}}liobait{\.e}, Bifet, Pechenizkiy, and
  Bouchachia]{gama2014survey}
J.~Gama, I.~{\v{Z}}liobait{\.e}, A.~Bifet, M.~Pechenizkiy, and A.~Bouchachia.
\newblock A survey on concept drift adaptation.
\newblock \emph{ACM computing surveys}, 46\penalty0 (4):\penalty0 1--37, 2014.

\bibitem[Giritlio{\u{g}}lu et~al.(2021)Giritlio{\u{g}}lu, Mandira, Yilmaz,
  Ertenli, Akg{\"u}r, K{\i}n{\i}kl{\i}o{\u{g}}lu, Kurt, Mutlu, G{\"u}rel, and
  Dibeklio{\u{g}}lu]{giritliouglu2021multimodal}
D.~Giritlio{\u{g}}lu, B.~Mandira, S.~F. Yilmaz, C.~U. Ertenli, B.~F. Akg{\"u}r,
  M.~K{\i}n{\i}kl{\i}o{\u{g}}lu, A.~G. Kurt, E.~Mutlu, {\c{S}}.~C. G{\"u}rel,
  and H.~Dibeklio{\u{g}}lu.
\newblock Multimodal analysis of personality traits on videos of
  self-presentation and induced behavior.
\newblock \emph{Journal on Multimodal User Interfaces}, 15\penalty0
  (4):\penalty0 337--358, 2021.

\bibitem[Gomes et~al.(2025)Gomes, Lee, Gunasekara, Sun, Cassales, Liu, Heyden,
  Cerqueira, Bahri, Koh, Pfahringer, and
  Bifet]{gomes2025capymoaefficientmachinelearning}
H.~M. Gomes, A.~Lee, N.~Gunasekara, Y.~Sun, G.~W. Cassales, J.~J. Liu,
  M.~Heyden, V.~Cerqueira, M.~Bahri, Y.~S. Koh, B.~Pfahringer, and A.~Bifet.
\newblock {CapyMOA}: Efficient machine learning for data streams in python,
  2025.
\newblock URL \url{https://arxiv.org/abs/2502.07432}.

\bibitem[{H}enzinger et~al.(1998){H}enzinger, {R}aghavan, and
  {R}ajagopalan]{henzinger1998computing}
M.~R. {H}enzinger, P.~{R}aghavan, and S.~{R}ajagopalan.
\newblock {C}omputing on data streams.
\newblock \emph{{E}xternal {M}emory {A}lgorithms}, 50:\penalty0 107--118, 1998.

\bibitem[{H}ido et~al.(2013){H}ido, {T}okui, and {O}da]{hido2013jubatus}
S.~{H}ido, S.~{T}okui, and S.~{O}da.
\newblock {J}ubatus: {A}n open source platform for distributed online machine
  learning.
\newblock In \emph{{N}{I}{P}{S} 2013 {W}orkshop on {B}ig {L}earning, {L}ake
  {T}ahoe}, 2013.

\bibitem[{I}shimtsev et~al.(2017){I}shimtsev, {B}ernstein, {B}urnaev, and
  {N}azarov]{ishimtsev2017conformal}
V.~{I}shimtsev, A.~{B}ernstein, E.~{B}urnaev, and I.~{N}azarov.
\newblock {C}onformal $ k $-nn anomaly detector for univariate data streams.
\newblock In \emph{{C}onformal and {P}robabilistic {P}rediction and
  {A}pplications}, pages 213--227, 2017.

\bibitem[{K}loft and {L}askov(2010)]{kloft2010online}
M.~{K}loft and P.~{L}askov.
\newblock {O}nline anomaly detection under adversarial impact.
\newblock In \emph{{P}roceedings of the {T}hirteenth {I}nternational
  {C}onference on {A}rtificial {I}ntelligence and {S}tatistics}, pages
  405--412, 2010.

\bibitem[{L}e and {H}o(2005)]{le2005association}
S.~Q. {L}e and T.~B. {H}o.
\newblock {A}n association-based dissimilarity measure for categorical data.
\newblock \emph{{P}attern {R}ecognition {L}etters}, 26\penalty0 (16):\penalty0
  2549--2557, 2005.

\bibitem[Mand{\i}ra et~al.(2019)Mand{\i}ra, Giritlioglu, Yilmaz, Ertenli,
  Akg{\"u}r, K{\i}n{\i}kl{\i}o{\u{g}}lu, Kurt, Doganl{\i}, Mutlu, G{\"u}rel,
  et~al.]{mandira2019spatiotemporal}
B.~Mand{\i}ra, D.~Giritlioglu, S.~F. Yilmaz, C.~U. Ertenli, B.~F. Akg{\"u}r,
  M.~K{\i}n{\i}kl{\i}o{\u{g}}lu, A.~G. Kurt, M.~N. Doganl{\i}, E.~Mutlu, S.~C.
  G{\"u}rel, et~al.
\newblock Spatiotemporal and multimodal analysis of personality traits.
\newblock In \emph{15th International Summer Workshop on Multimodal
  Interfaces}, page~32, 2019.

\bibitem[Manzoor et~al.(2018)Manzoor, Lamba, and Akoglu]{manzoor2018xstream}
E.~Manzoor, H.~Lamba, and L.~Akoglu.
\newblock xstream: Outlier detection in feature-evolving data streams.
\newblock In \emph{Proceedings of the 24th ACM SIGKDD International Conference
  on Knowledge Discovery \& Data Mining}, pages 1963--1972, 2018.

\bibitem[{M}ontiel et~al.(2018){M}ontiel, {R}ead, {B}ifet, and
  {A}bdessalem]{montiel2018scikit}
J.~{M}ontiel, J.~{R}ead, A.~{B}ifet, and T.~{A}bdessalem.
\newblock {S}cikit-multiflow: {A} multi-output streaming framework.
\newblock \emph{{J}ournal of {M}achine {L}earning {R}esearch}, 19\penalty0
  (1):\penalty0 2915--2914, 2018.

\bibitem[Montiel et~al.(2021)Montiel, Halford, Mastelini, Bolmier, Sourty,
  Vaysse, Zouitine, Gomes, Read, Abdessalem, et~al.]{montiel2021river}
J.~Montiel, M.~Halford, S.~M. Mastelini, G.~Bolmier, R.~Sourty, R.~Vaysse,
  A.~Zouitine, H.~M. Gomes, J.~Read, T.~Abdessalem, et~al.
\newblock River: machine learning for streaming data in python.
\newblock \emph{Journal of Machine Learning Research}, 22\penalty0
  (110):\penalty0 1--8, 2021.

\bibitem[Pedregosa et~al.(2011)Pedregosa, Varoquaux, Gramfort, Michel, Thirion,
  Grisel, Blondel, Prettenhofer, Weiss, Dubourg, et~al.]{sklearn}
F.~Pedregosa, G.~Varoquaux, A.~Gramfort, V.~Michel, B.~Thirion, O.~Grisel,
  M.~Blondel, P.~Prettenhofer, R.~Weiss, V.~Dubourg, et~al.
\newblock Scikit-learn: Machine learning in python.
\newblock \emph{Journal of machine learning research}, 12:\penalty0 2825--2830,
  2011.

\bibitem[Pevn{\'y}(2016)]{pevny2016loda}
T.~Pevn{\'y}.
\newblock Loda: Lightweight on-line detector of anomalies.
\newblock \emph{Machine Learning}, 102\penalty0 (2):\penalty0 275--304, 2016.

\bibitem[{R}en et~al.(2019){R}en, {X}u, {W}ang, Yi, {H}uang, {K}ou, {X}ing,
  Yang, {T}ong, and {Z}hang]{ren2019time}
H.~{R}en, B.~{X}u, Y.~{W}ang, C.~Yi, C.~{H}uang, X.~{K}ou, T.~{X}ing, M.~Yang,
  J.~{T}ong, and Q.~{Z}hang.
\newblock {T}ime-series anomaly detection service at microsoft.
\newblock In \emph{{P}roceedings of the 25th {A}{C}{M} {S}{I}{G}{K}{D}{D}
  {I}nternational {C}onference on {K}nowledge {D}iscovery \& {D}ata {M}ining},
  pages 3009--3017, 2019.

\bibitem[{S}afin and {B}urnaev(2017)]{safin2017conformal}
A.~M. {S}afin and E.~{B}urnaev.
\newblock {C}onformal kernel expected similarity for anomaly detection in
  time-series data.
\newblock \emph{{A}dvances in {S}ystems {S}cience and {A}pplications},
  17\penalty0 (3):\penalty0 22--33, 2017.

\bibitem[{T}an et~al.(2011){T}an, {T}ing, and {L}iu]{tan2011fast}
S.~C. {T}an, K.~M. {T}ing, and T.~F. {L}iu.
\newblock {F}ast anomaly detection for streaming data.
\newblock In \emph{{T}wenty-{S}econd {I}nternational {J}oint {C}onference on
  {A}rtificial {I}ntelligence}, 2011.

\bibitem[Van Der~Walt et~al.(2011)Van Der~Walt, Colbert, and
  Varoquaux]{walt2011numpy}
S.~Van Der~Walt, S.~C. Colbert, and G.~Varoquaux.
\newblock The {NumPy} array: a structure for efficient numerical computation.
\newblock \emph{Computing in science \& engineering}, 13\penalty0 (2):\penalty0
  22--30, 2011.

\bibitem[Virtanen et~al.(2020)Virtanen, Gommers, Oliphant, Haberland, Reddy,
  Cournapeau, Burovski, Peterson, Weckesser, Bright, et~al.]{virtanen2020scipy}
P.~Virtanen, R.~Gommers, T.~E. Oliphant, M.~Haberland, T.~Reddy, D.~Cournapeau,
  E.~Burovski, P.~Peterson, W.~Weckesser, J.~Bright, et~al.
\newblock {SciPy} 1.0: fundamental algorithms for scientific computing in
  {Python}.
\newblock \emph{Nature methods}, 17\penalty0 (3):\penalty0 261--272, 2020.

\bibitem[{W}en(2020)]{adtk}
T.~{W}en.
\newblock {A}dtk: {A}nomaly detection toolkit, 2020.
\newblock URL \url{https://github.com/arundo/adtk}.

\bibitem[Y{\i}lmaz(2021)]{yilmaz2021unsupervised}
S.~F. Y{\i}lmaz.
\newblock Unsupervised anomaly detection via deep metric learning with
  end-to-end optimization.
\newblock Master's thesis, Bilkent University (T{\"u}rkiye), 2021.

\bibitem[Yilmaz and Kozat(2020{\natexlab{a}})]{yilmaz2020face}
S.~F. Yilmaz and S.~S. Kozat.
\newblock Face presentation attack detection via spatiotemporal autoencoder.
\newblock In \emph{IEEE Signal Processing and Communications Applications
  Conference}, 2020{\natexlab{a}}.

\bibitem[Yilmaz and Kozat(2020{\natexlab{b}})]{yilmaz2020robust}
S.~F. Yilmaz and S.~S. Kozat.
\newblock Robust anomaly detection via sequential ensemble learning.
\newblock In \emph{IEEE Signal Processing and Communications Applications
  Conference}, 2020{\natexlab{b}}.

\bibitem[Yuan et~al.(2014)Yuan, {F}ang, and {W}ang]{yuan2014online}
Y.~Yuan, J.~{F}ang, and Q.~{W}ang.
\newblock {O}nline anomaly detection in crowd scenes via structure analysis.
\newblock \emph{{I}{E}{E}{E} {T}ransactions on {C}ybernetics}, 45\penalty0
  (3):\penalty0 548--561, 2014.

\bibitem[{Z}hao et~al.(2019){Z}hao, {Z}ain {N}asrullah, and {Z}heng {L}i]{pyod}
Y.~{Z}hao, {Z}ain {N}asrullah, and {Z}heng {L}i.
\newblock {P}yod: {A} python toolbox for scalable outlier detection.
\newblock \emph{{J}ournal of {M}achine {L}earning {R}esearch}, 20\penalty0
  (96):\penalty0 1--7, 2019.

\end{thebibliography}

\end{document}